\documentclass[runningheads]{llncs}
\usepackage{amsmath}
\usepackage{amsfonts}
\usepackage[T1]{fontenc}
\usepackage{float}

\usepackage{graphicx}
\begin{document}
\title{Decomposition-Guided Curvelet Thresholding for Sharp-to-Soft CT Kernel Conversion}
%
%
\author{Mahmoud Nasr\inst{1,2}\and
Jan K. Argasiński\inst{2,3}\and
Krzysztof Brzostowski\inst{4} \and Adam Piórkowski\inst{1}
}
\authorrunning{M. Nasr et al.}
%
\institute{
Department of Biocybernetics and Biomedical Engineering, AGH University of Krakow, 30-059 Krakow, Poland. \and Sano Centre for Computational Medicine, Czarnowiejska 36/C5, Kraków, 30-054, Poland. \and Faculty of Physics, Astronomy and Applied Computer Science, Jagiellonian University, Łojasiewicza 11, Kraków, 30-348, Poland. \and
Faculty of Information and Communication Technology, Wroclaw University of Science and Technology,  Wyb. Wyspia\'{n}skiego 27, 50-370 Wroclaw,
 Poland.
\\
}
\maketitle              
\begin{abstract}

Image denoising is a crucial task in image processing, focused on improving image quality by minimizing noise while maintaining essential structural elements. This study presents a hybrid denoising framework that combines several decomposition techniques, including empirical mode decomposition (EMD), variational mode decomposition (VMD), multichannel EMD (MEMD), and bidimensional EMD (BEMD), with curvelet transform thresholding. Each decomposition mode undergoes processing through both soft and hard thresholding, and the denoised modes are combined to rebuild the final image. Comprehensive evaluations of standard CT image datasets reconstructed with various kernels (B50, B46, B41, B36) reveal substantial enhancements in denoising efficacy. VMD consistently achieves the highest peak signal-to-noise ratio (PSNR) and structural similarity index (SSIM), signifying exceptional noise reduction and feature preservation. The study analyses the trade-offs between soft and hard thresholding: soft thresholding maintains intricate visual details, whilst harsh thresholding provides enhanced noise reduction. The suggested method surpasses traditional techniques in both reference and non-reference quality criteria, indicating its potential for broader application in medical imaging and future incorporation with adaptive thresholding algorithms.

\end{abstract}
\section{Introduction}
Image denoising is a fundamental operation in image processing. Its goal is to improve image quality by eliminating extraneous noise while maintaining critical characteristics. This method is essential for numerous applications, including medical imaging, remote sensing, and digital photography, where noise can severely impair visual quality and the precision of subsequent image analysis activities \cite{fan2019brief,sagheer2020review,nasr2025comparative}.

Throughout the years, various solutions have been devised to tackle the issue of image denoising. Decomposition-based approaches have garnered significant interest because of their ability to differentiate noise from the fundamental information of the image adaptively. Empirical Mode Decomposition (EMD) and its variations \cite{boudraa2007emd,linderhed2009image}, including variational mode decomposition (VMD) \cite{lahmiri2014biomedical,brzostowski2024combining}, multichannel empirical mode decomposition (MEMD) \cite{fleureau2011multivariate}, and bidimensional empirical mode decomposition (BEMD) \cite{nunes2003image}, have demonstrated efficacy in this context. These techniques convert an image into a collection of intrinsic mode functions (IMFs) or bidimensional intrinsic mode functions (BIMFs), facilitating precise noise removal \cite{liu2019image}.


In addition to decomposition approaches, multiscale transforms have demonstrated significant potential in image denoising. The Curvelet Transform provides substantial benefits in detecting image edges and intricate details, making it appropriate for denoising \cite{Starck2002,KaurBains2015,Neema2016}. Utilizing thresholding techniques, including soft and hard thresholding \cite{seelaboyina2023different,jardim2023image}, on the curvelet coefficients enables excellent noise suppression while maintaining crucial image information \cite{yulong2023wavelets}.

This work presents an innovative method for image denoising that integrates decomposition modes with Curvelet Transform thresholding. The image is initially broken into various modes using EMD, VMD, Multichannel EMD, and BEMD. Each deconstructed mode undergoes Curvelet Transform thresholding, employing both soft and hard thresholding approaches \cite{biswas2012new,lei2021remote}. The denoised modes are then aggregated to reconstitute the final denoised image.

The efficacy of the suggested strategy is assessed by comprehensive tests on standard image datasets. The evaluation of performance employs both reference metrics, including Peak Signal-to-Noise Ratio (PSNR), Signal-to-Noise Ratio (SNR), and Structural Similarity Index (SSIM), as well as non-reference metrics, such as Natural Image Quality Evaluator (NIQE) Score, Blind/Referenceless Image Spatial Quality Evaluator (BRISQUE) Score, Perceptual Image Quality Evaluator (PIQE) Score, and Entropy. The results illustrate the suggested method's capacity to effectuate substantial noise reduction while preserving the integrity of image features.

\section{Background}

Image denoising aims to remove extraneous noise while preserving critical image characteristics. Among numerous techniques, Empirical Mode Decomposition (EMD) and its derivatives have garnered considerable interest.

\subsection{Empirical Mode Decomposition (EMD)}

Empirical Mode Decomposition (EMD) is a data-driven methodology that decomposes signals into intrinsic mode functions (IMFs) through iterative sifting processes. The procedure entails finding local extrema, interpolating them to generate envelopes, averaging the envelopes, and then subtracting this mean from the signal to produce a proto-IMF. This persists until an IMF is acquired. The IMF is subsequently eliminated from the signal, and the procedure is reiterated on the residual until complete decomposition is attained \cite{huang1998empirical,rilling2003empirical,torres2011complete}.

The final result is a set of IMFs and a residual capturing the signal's non-oscillatory trend:

\begin{equation}
    x(t) = \sum_{i=1}^n c_i(t) + r(t)
\end{equation}

Empirical Mode Decomposition (EMD) is adept at handling nonlinear, nonstationary data and has demonstrated efficacy in image denoising by distinguishing noise from structure elements \cite{boudraa2007emd,linderhed2009image}. To mitigate the limitations of EMD, including mode mixing, upgrades such as Ensemble EMD (EEMD) \cite{wu2009ensemble} and Complete Ensemble EMD (CEEMD) \cite{ali2020complete} have been proposed, including noise-assisted techniques to increase robustness.

\subsection{Variational Mode Decomposition (VMD)}

In contrast to EMD, VMD employs a variational method to deconstruct a signal into a specified number of modes by resolving an optimisation problem:

\begin{equation}
\{u_k\}, \{\omega_k\} = \arg\min_{\{u_k\}, \{\omega_k\}} \left\{ \sum_{k} \left\| \partial_t \left[ \left( \delta(t) + \frac{j}{\pi t} \right) * u_k(t) \right] e^{-j \omega_k t} \right\|_2^2 \right\}
\end{equation}

This approach guarantees that modes are distinctly separated in the frequency domain, hence reducing mode mixing and improving decomposition precision.

Both EMD and VMD demonstrate the functionality of the filter bank, facilitating the efficient segregation of frequency components, essential for reducing image noise \cite{lahmiri2014biomedical,liu2019image,qi2022method}.

\subsection{Multichannel Empirical Mode Decomposition (MEMD)}

MEMD expands EMD to multichannel signals, facilitating the concurrent decomposition of associated signals, such as RGB channels in color images:

\begin{equation}
    x_i(t) = \sum_{j=1}^n c_{i,j}(t) + r_i(t)
\end{equation}

This joint processing helps preserve inter-channel correlations during denoising.

\subsubsection{Bidimensional Empirical Mode Decomposition (BEMD)}
BEMD is specifically designed for image processing and decomposes the image into bidimensional intrinsic mode functions (BIMFs). The procedure entails examining the image in two dimensions to derive BIMFs, which signify various frequency components of the image.

\subsection{Curvelet Transform}

The Curvelet Transform is a multiscale, multidirectional technique that surpasses wavelets in detecting image edges and singularities. The process commences with grayscale conversion and normalization, followed by a two-dimensional Fast Fourier Transform (FFT). Directional frequency windows delineate characteristics at diverse sizes and orientations. The filtered coefficients are subsequently transformed back using the inverse FFT to acquire curvelet coefficients.

Noise is reduced through thresholding—either soft:

\begin{equation}
    \hat{C}_{\text{soft}}[f](a,b,\theta) = \text{sign}(C[f](a,b,\theta)) \cdot \max(|C[f](a,b,\theta)| - \lambda, 0)
\end{equation}

or hard:

\begin{equation}
\hat{C}_{\text{hard}}[f](a,b,\theta) = 
\begin{cases} 
C[f](a,b,\theta), & \text{if } |C[f](a,b,\theta)| > \lambda \\
0, & \text{otherwise}
\end{cases}
\end{equation}

Following thresholding, the inverse Curvelet Transform reconstructs the denoised image, preserving essential characteristics while reducing noise.

\section{Methodology}
\begin{figure}[ht!]
\centering
\includegraphics[width=0.9\textwidth]{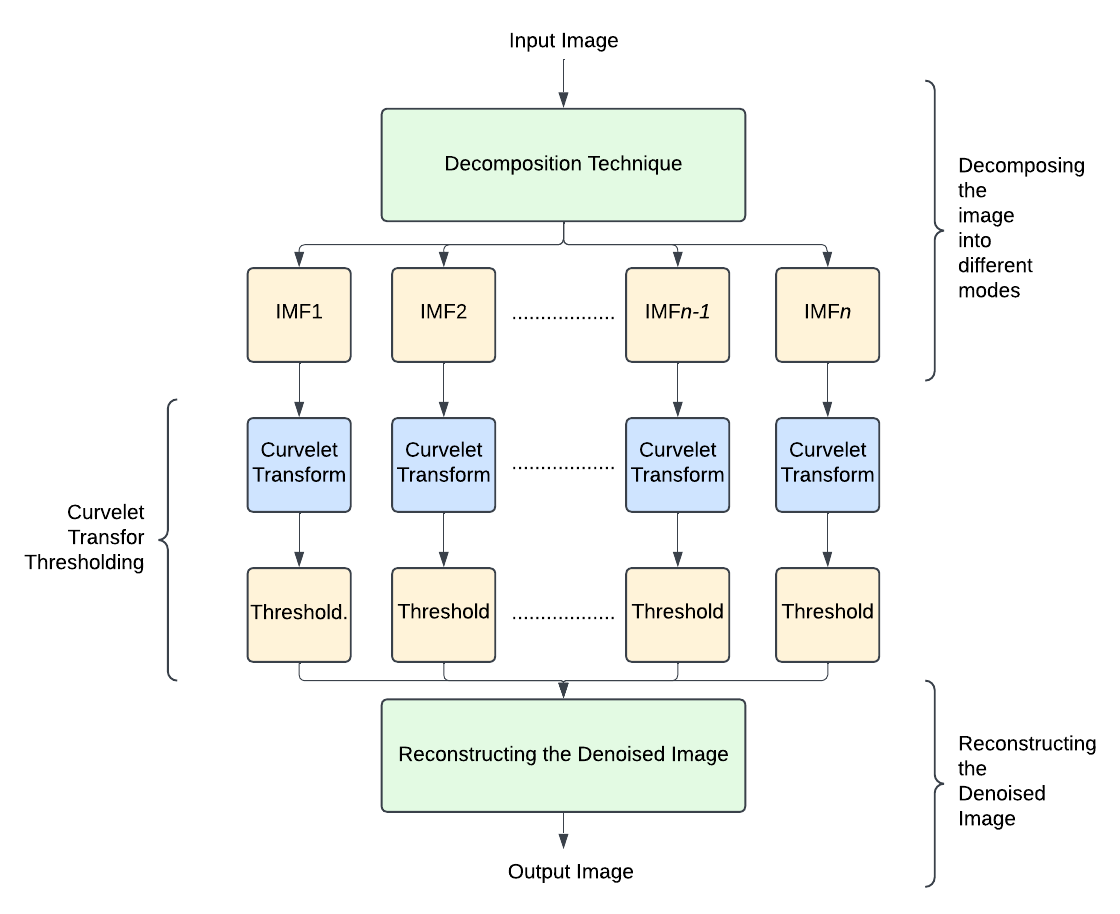}
\caption{This figure shows the structure of the proposed method based on the 3 steps for image denoising and enhancing image quality.} \label{fig1}
\end{figure}

The methods section delineates the procedures employed for image denoising using decomposition modes utilizing curvelet transform thresholding. The procedure consists of three primary phases, as shown in Figure \ref{fig1}: disaggregating the image into several modes, implementing curvelet transform thresholding, and constructing the denoised image after thresholding.

As discussed in Section 2, we utilize decomposition mode techniques alongside the Curvelet Transform and thresholding methods to perform image denoising. The method comprises three main phases. Initially, the chosen decomposition mode is applied to the input image, resulting in a set of Intrinsic Mode Functions (IMFs). In the second phase, the Curvelet Transform is applied to each IMF separately, followed by thresholding using either soft or hard techniques. The thresholded IMFs are then obtained by applying the inverse Curvelet Transform. In the final phase, the denoised image is reconstructed by summing all the thresholded IMFs and residual, which are subsequently evaluated for performance.

\section{Materials and Evaluation Methods}

\subsection{Datasets}

This study uses Coronary Computed Tomography Angiography (CCTA) scans acquired using a Siemens SOMATOM Definition Flash scanner. The dataset used includes 10 reconstructed CT images with a range of analytical reconstruction kernels: B30f (smooth), B31f, B35f, B36f (medium soft), B41f, B46f (medium), and B50f (medium sharp) \cite{arcadi2015coronary,ciolek2020image}. These "B" series kernels are standard in body imaging. Among the kernels, B30f, known for its high contrast-to-noise ratio (CNR), was selected as the reference standard. All images were provided in 12-bit DICOM format \cite{lasek2021ct,nasr2023approach,nasr2025novel}.

\subsection{Implementation}

All decomposition and denoising procedures were implemented in MATLAB. The workflow began with image decomposition using EMD, VMD, Multichannel EMD, and BEMD. Each resulting mode was then processed using the curvelet transform, followed by noise reduction through soft and hard thresholding. Finally, the denoised image was reconstructed by summing the thresholded IMFs or BIMFs, producing a noise-suppressed output.

\subsection{Evaluation Metrics}

To evaluate denoising performance, both full-reference and no-reference image quality metrics were used. The full-reference metrics included Peak Signal-to-Noise Ratio (PSNR), Signal-to-Noise Ratio (SNR), and Structural Similarity Index Measure (SSIM), which compare the denoised images to the original ground truth \cite{hore2010image,jagalingam2015review,bielecka2020universal}. For blind evaluation, non-reference metrics such as NIQE, BRISQUE, PIQE, and image entropy were applied \cite{herzog2012norm,woodard2006no}. These scores offered insights into image quality when a clean reference was unavailable. The results from all metrics were systematically analyzed to compare the performance of the denoising methods.




\section{Results and Discussion}

This section presents the denoising results obtained using the proposed techniques and evaluates their effectiveness based on both reference and no-reference image quality metrics. These include PSNR, SNR, SSIM (reference-based), and NIQE, BRISQUE, PIQE, and entropy (no-reference).

\subsection{Soft Thresholding Results}

\begin{figure}[ht!]
\centering
\includegraphics[width=0.7\textwidth]{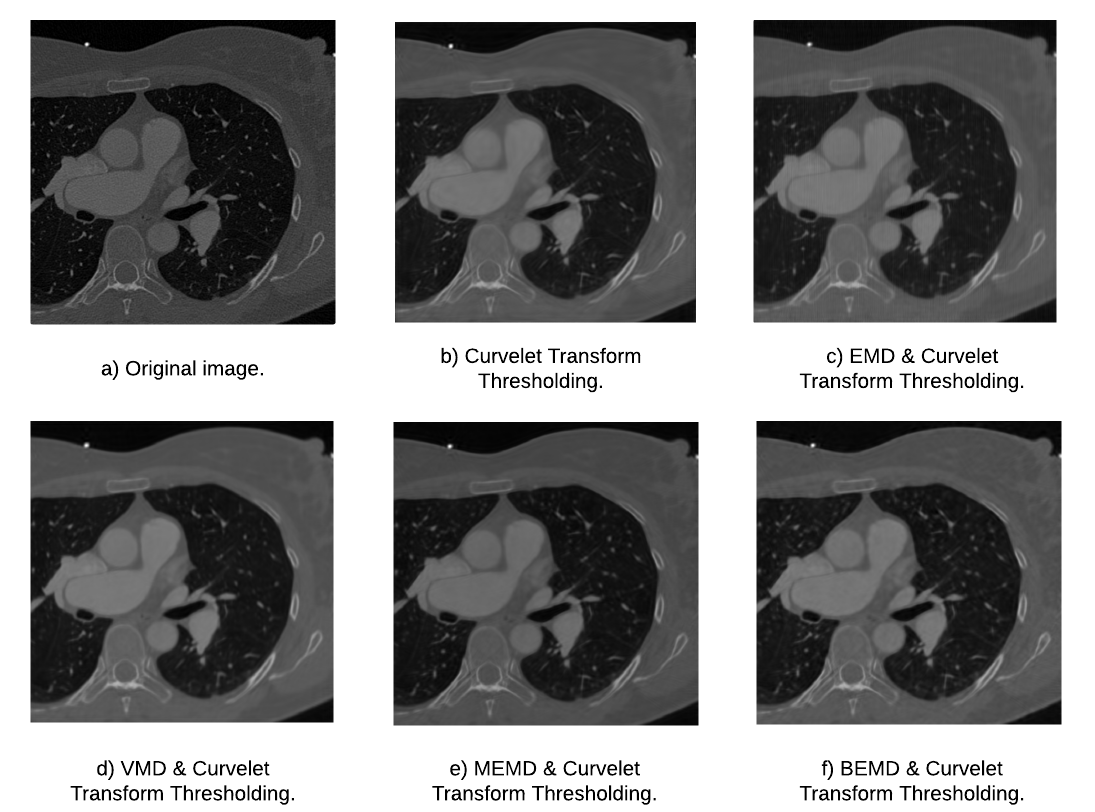}
\caption{Samples of the denoised images of an input reconstructed image B50 (sharp Kernel) by applying the soft threshold.} \label{fig2}
\end{figure}

Denoising results using soft thresholding for selected CT reconstruction kernels (B50, B46, B41, B36) are illustrated in Figure~\ref{fig2}, with corresponding performance metrics summarized in Tables~\ref{tab:soft_thresholding_B50} through \ref{tab:soft_thresholding_B36}. Each table provides a side-by-side comparison of decomposition methods: standard Curvelet thresholding, EMD-Curvelet, VMD-Curvelet, MEMD-Curvelet, and BEMD-Curvelet.

\begin{table}[ht!] \centering \caption{Performance Metrics for Soft Thresholding (B50)} \label{tab:soft_thresholding_B50}
\resizebox{\columnwidth}{!}{%
\begin{tabular}{|c|c|c|c|c|c|c|c|} \hline \textbf{Metric} & \textbf{Curvelet} & \textbf{EMD + Curvelet} & \textbf{VMD + Curvelet} & \textbf{MEMD + Curvelet} & \textbf{BEMD + Curvelet} \\ \hline SSIM & 0.931 & 0.903 & 0.949 & 0.954 & 0.949 \\ \hline PSNR & 32.344 & 32.548 & 33.894 & 33.094 & 32.816 \\ \hline SNR & 19.079 & 19.283 & 20.629 & 19.829 & 19.551 \\ \hline NIQE Score & 4.656 & 9.543 & 5.396 & 5.040 & 5.056 \\ \hline BRISQUE Score & 40.575 & 42.032 & 44.369 & 42.191 & 42.331 \\ \hline PIQE Score & 65.059 & 66.000 & 61.523 & 75.042 & 70.478 \\ \hline Entropy & 5.970 & 6.097 & 6.018 & 5.955 & 6.022 \\ \hline \end{tabular}} \end{table}

\begin{table}[ht!] \centering \caption{Performance Metrics for Soft Thresholding (B46)} \label{tab:soft_thresholding_B46}
\resizebox{\columnwidth}{!}{%
\begin{tabular}{|c|c|c|c|c|c|c|c|} \hline \textbf{Metric} & \textbf{Curvelet} & \textbf{EMD + Curvelet} & \textbf{VMD + Curvelet} & \textbf{MEMD + Curvelet} & \textbf{BEMD + Curvelet} \\ \hline SSIM & 0.949 & 0.915 & 0.954 & 0.957 & 0.953 \\ \hline PSNR & 37.856 & 36.516 & 38.475 & 38.187 & 37.796 \\ \hline SNR & 24.523 & 23.183 & 25.143 & 24.854 & 24.463 \\ \hline NIQE Score & 4.686 & 9.888 & 5.061 & 5.294 & 5.340 \\ \hline BRISQUE Score & 42.536 & 41.963 & 40.965 & 43.431 & 43.758 \\ \hline PIQE Score & 71.384 & 71.945 & 73.732 & 76.598 & 71.873 \\ \hline Entropy & 5.912 & 6.054 & 5.959 & 5.897 & 5.941 \\ \hline \end{tabular} }\end{table}

\begin{table}[ht!] \centering \caption{Performance Metrics for Soft Thresholding (B41)} \label{tab:soft_thresholding_B41} 
\resizebox{\columnwidth}{!}{%
\begin{tabular}{|c|c|c|c|c|c|c|c|} \hline \textbf{Metric} & \textbf{Curvelet} & \textbf{EMD + Curvelet} & \textbf{VMD + Curvelet} & \textbf{MEMD + Curvelet} & \textbf{BEMD + Curvelet} \\ \hline SSIM & 0.949 & 0.913 & 0.957 & 0.958 & 0.953 \\ \hline PSNR & 38.515 & 37.009 & 39.528 & 39.358 & 38.803 \\ \hline SNR & 26.551 & 25.044 & 27.564 & 27.394 & 26.839 \\ \hline NIQE Score & 4.937 & 11.547 & 5.510 & 5.515 & 5.537 \\ \hline BRISQUE Score & 43.902 & 42.462 & 43.255 & 43.603 & 43.923 \\ \hline PIQE Score & 78.073 & 75.470 & 77.409 & 80.670 & 78.187 \\ \hline Entropy & 6.136 & 6.273 & 6.162 & 6.112 & 6.149 \\ \hline \end{tabular}}\end{table}

\begin{table}[ht!] \centering \caption{Performance Metrics for Soft Thresholding (B36)} \label{tab:soft_thresholding_B36}
\resizebox{\columnwidth}{!}{%
\begin{tabular}{|c|c|c|c|c|c|c|c|} \hline \textbf{Metric} & \textbf{Curvelet} & \textbf{EMD + Curvelet} & \textbf{VMD + Curvelet} & \textbf{MEMD + Curvelet} & \textbf{BEMD + Curvelet} \\ \hline SSIM & 0.964 & 0.915 & 0.957 & 0.959 & 0.958 \\ \hline PSNR & 43.112 & 38.321 & 41.343 & 41.700 & 41.550 \\ \hline SNR & 30.687 & 25.896 & 28.917 & 29.274 & 29.124 \\ \hline NIQE Score & 5.018 & 11.728 & 5.171 & 5.322 & 5.226 \\ \hline BRISQUE Score & 42.755 & 42.630 & 44.310 & 44.577 & 44.366 \\ \hline PIQE Score & 78.889 & 75.834 & 80.792 & 83.809 & 82.571 \\ \hline Entropy & 5.995 & 6.170 & 6.034 & 5.991 & 6.017 \\ \hline \end{tabular} }\end{table}


The evaluation reveals that VMD consistently delivers the best denoising results, showing the highest PSNR and SSIM across most kernel types. These figures suggest superior noise suppression with minimal loss of structural detail. The improvement is further supported by no-reference metrics, where VMD yields lower NIQE, BRISQUE, and PIQE scores, reflecting better visual quality and reduced distortion.



\subsection{Hard Thresholding Results}

\begin{figure}[ht!]
\centering
\includegraphics[width=0.7\textwidth]{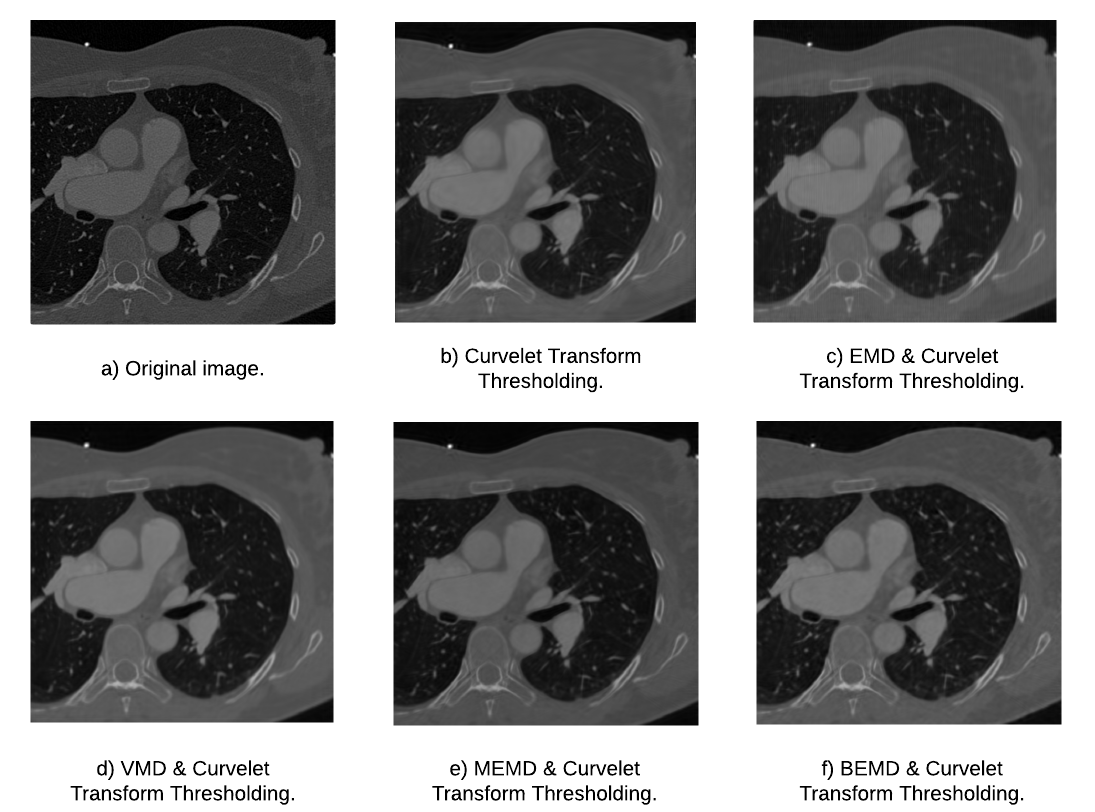}
\caption{Samples of the denoised images of an input reconstructed image B50 (sharp Kernel) by applying the hard threshold.} \label{fig3}
\end{figure}

Figure~\ref{fig3} and Tables~\ref{tab:hard_thresholding_B50} to \ref{tab:hard_thresholding_B36} present the denoising results using hard thresholding. Similar to the soft thresholding case, a comparative analysis is shown across the same set of decomposition strategies.

\begin{table}[ht!] 
\centering 
\caption{Performance Metrics for Hard Thresholding (B50)} \label{tab:hard_thresholding_B50}
\resizebox{\columnwidth}{!}{%
\begin{tabular}{|c|c|c|c|c|c|c|c|} \hline \textbf{Metric} & \textbf{Curvelet} & \textbf{EMD + Curvelet} & \textbf{VMD + Curvelet} & \textbf{MEMD + Curvelet} & \textbf{BEMD + Curvelet} \\ \hline SSIM & 0.945 & 0.900 & 0.940 & 0.952 & 0.946 \\ \hline PSNR & 35.905 & 35.650 & 37.429 & 35.909 & 35.731 \\ \hline SNR & 22.640 & 22.385 & 24.164 & 22.643 & 22.466 \\ \hline NIQE Score & 4.486 & 7.464 & 5.509 & 5.117 & 5.057 \\ \hline BRISQUE Score & 42.684 & 36.714 & 39.820 & 40.952 & 44.160 \\ \hline PIQE Score & 46.854 & 26.838 & 36.184 & 48.065 & 49.961 \\ \hline Entropy & 6.062 & 6.179 & 6.131 & 6.040 & 6.095 \\ \hline \end{tabular} }\end{table}

\begin{table}[ht!] 
\centering 
\caption{Performance Metrics for Hard Thresholding (B46)} \label{tab:hard_thresholding_B46}
\resizebox{\columnwidth}{!}{%
\begin{tabular}{|c|c|c|c|c|c|c|c|} \hline \textbf{Metric} & \textbf{Curvelet} & \textbf{EMD + Curvelet} & \textbf{VMD + Curvelet} & \textbf{MEMD + Curvelet} & \textbf{BEMD + Curvelet} \\ \hline SSIM & 0.971 & 0.941 & 0.972 & 0.972 & 0.970 \\ \hline PSNR & 42.329 & 40.082 & 42.587 & 41.754 & 41.914 \\ \hline SNR & 28.997 & 26.750 & 29.254 & 28.421 & 28.581 \\ \hline NIQE Score & 4.708 & 7.997 & 4.785 & 5.184 & 5.056 \\ \hline BRISQUE Score & 40.812 & 38.801 & 45.586 & 39.776 & 40.756 \\ \hline PIQE Score & 59.926 & 38.451 & 52.565 & 65.817 & 62.978 \\ \hline Entropy & 5.951 & 6.064 & 5.984 & 5.918 & 5.956 \\ \hline \end{tabular} }\end{table}

\begin{table}[ht!] 
\centering 
\caption{Performance Metrics for Hard Thresholding (B41)} \label{tab:hard_thresholding_B41} 
\resizebox{\columnwidth}{!}{%
\begin{tabular}{|c|c|c|c|c|c|c|c|} \hline \textbf{Metric} & \textbf{Curvelet} & \textbf{EMD + Curvelet} & \textbf{VMD + Curvelet} & \textbf{MEMD + Curvelet} & \textbf{BEMD + Curvelet} \\ \hline SSIM & 0.975 & 0.949 & 0.978 & 0.977 & 0.977 \\ \hline PSNR & 42.930 & 40.783 & 43.812 & 43.178 & 43.184 \\ \hline SNR & 30.966 & 28.819 & 31.848 & 31.214 & 31.220 \\ \hline NIQE Score & 5.114 & 8.948 & 5.128 & 5.582 & 5.387 \\ \hline BRISQUE Score & 45.910 & 41.408 & 47.861 & 43.284 & 44.929 \\ \hline PIQE Score & 66.931 & 46.319 & 53.507 & 75.715 & 69.959 \\ \hline Entropy & 6.157 & 6.263 & 6.187 & 6.129 & 6.156 \\ \hline \end{tabular} }\end{table}

\begin{table}[ht!] 
\centering 
\caption{Performance Metrics for Hard Thresholding (B36)} \label{tab:hard_thresholding_B36}
\resizebox{\columnwidth}{!}{%
\begin{tabular}{|c|c|c|c|c|c|c|c|} \hline \textbf{Metric} & \textbf{Curvelet} & \textbf{EMD + Curvelet} & \textbf{VMD + Curvelet} & \textbf{MEMD + Curvelet} & \textbf{BEMD + Curvelet} \\ \hline SSIM & 0.979 & 0.947 & 0.974 & 0.974 & 0.975 \\ \hline PSNR & 48.418 & 42.252 & 45.942 & 46.410 & 46.836 \\ \hline SNR & 35.992 & 29.827 & 33.516 & 33.984 & 34.410 \\ \hline NIQE Score & 5.308 & 9.469 & 5.020 & 5.348 & 5.172 \\ \hline BRISQUE Score & 40.779 & 40.778 & 40.380 & 42.512 & 41.557 \\ \hline PIQE Score & 67.228 & 51.531 & 69.299 & 77.144 & 76.257 \\ \hline Entropy & 5.999 & 6.134 & 6.027 & 5.979 & 5.999 \\ \hline \end{tabular}} \end{table}


The results reaffirm VMD’s effectiveness, as it again achieves the top PSNR and SSIM scores for most kernels. However, while hard thresholding offers more aggressive noise reduction, it also introduces a slight loss of fine details, as evidenced by marginally higher NIQE and BRISQUE values. This suggests a trade-off: while structural clarity may improve, the subtle textural details can be slightly diminished.





\subsection{Discussion}

The evaluation findings indicate that VMD regularly exceeds alternative decomposition methods in both soft and hard thresholding contexts. The ability to separate specific frequency components enhances noise reduction and structural integrity, as evidenced by improved PSNR and SSIM scores and favorable perceptual metrics. Soft thresholding preserves more visual details, but harsh thresholding provides enhanced noise reduction, which may introduce artifacts.

The results indicate that the selection of decomposition approach and thresholding method should be determined by the particular needs of the application, prioritising either detail preservation or optimal noise suppression. The combination of decomposition and Curvelet-based thresholding effectively enhances image quality in CT datasets.




\section{Conclusion}

This study introduces a denoising framework that integrates decomposition-based signal separation with Curvelet Transform thresholding to enhance image quality in CT imaging. Adaptive decomposition approaches, particularly Variational Mode Decomposition (VMD), facilitate the efficient separation of noise components, while the Curvelet Transform excels at preserving directional and edge information during thresholding.

Experimental findings across various reconstruction kernels validate the method's efficacy, with VMD proving to be the most dependable approach. The juxtaposition of soft and hard thresholding highlights the need to balance noise attenuation and structural integrity, depending on diagnostic or visualization requirements.

Future research may focus on adaptive thresholding techniques and expand this methodology to additional medical imaging modalities, thereby facilitating enhanced practical applicability and real-time implementations.

\section*{Acknowledgements}

This project has received funding from the European Union's Horizon 2020 research and innovation programme under grant agreement No 857533 and from the International Research Agendas Programme of the Foundation for Polish Science No MAB PLUS/2019/13. 

The publication was created within the project of the Minister of Science and Higher Education "Support for the activity of Centers of Excellence established in Poland under Horizon 2020" on the basis of the contract number MEiN/2023/DIR/3796. 

Research project partly supported by the program "Excellence Initiative – Research University" for the AGH University of Krakow. 

The statistical analysis was conducted with the collaboration of the Regional Digital Medicine Centre at the 4th Military Clinical Hospital in Wrocław, Poland (Project No. 2023/ABM/02/00005-00; Medical Research Agency).
\bibliographystyle{unsrt}
\bibliography{ref}           

\end{document}